\documentclass{Interspeech2024}

\interspeechcameraready % Comment out for blind version

\usepackage{hyperref}

\title{Real-time Speech Summarization for Medical Conversations}

\name[affiliation={*1,2,6}]{Khai}{Le-Duc}
\name[affiliation={*3}]{Khai-Nguyen}{Nguyen}
\name[affiliation={4}]{Long}{Vo-Dang}
\name[affiliation={5,6}]{Truong-Son}{Hy}
\address{
  $^1$University of Toronto, Canada
  $^2$University Health Network, Canada \\
  $^3$College of William and Mary, United States
  $^4$University of Cincinnati, United States \\
  $^5$Indiana State University, United States
  $^6$FPT Software AI Center, Vietnam}
\email{duckhai.le@mail.utoronto.ca, knguyen07@wm.edu, TruongSon.Hy@indstate.edu}


\begin{document}

\maketitle

% \begin{abstract}
% % 1000 characters. ASCII characters only. No citations.
% Compared to previous design, our proposed design allows users to identify when a satisfactory summary is generated since the summary does not get updated every time a word is generated by ASR system but rather gets updated after N utterances, which can also potentially save computational cost.

\begin{abstract}
% In the course of real-world conversations, the amount of available information is rapidly increasing proportional to speaking speeds, causing information overload. 
% No one has the ability to remember every content in the conversation, but people often make decisions based on the degree of informational importance and critical information.

In doctor-patient conversations, identifying medically relevant information is crucial, posing the need for conversation summarization.
In this work, we propose the first deployable real-time speech summarization system for real-world applications in industry, which generates a local summary after every N speech utterances within a conversation and a global summary after the end of a conversation. 
Our system could enhance user experience from a business standpoint, while also reducing computational costs from a technical perspective.
Secondly, we present VietMed-Sum which, to our knowledge, is the first speech summarization dataset for medical conversations.
Thirdly, we are the first to utilize LLM and human annotators collaboratively to create gold standard and synthetic summaries for medical conversation summarization. 
Finally, we present baseline results of state-of-the-art models on VietMed-Sum.
All code, data (English-translated and Vietnamese) and models are available online.
\end{abstract}

\def\thefootnote{*}\footnotetext{Equal contribution}\def\thefootnote{\arabic{footnote}}

% Forced page numbering
\thispagestyle{plain}
\pagestyle{plain}

\section{Introduction}
% %\khai{DONE}

In real-world conversations, the volume of information grows significantly in tandem with speaking rates, leading to information overload. 
Remembering every detail discussed, especially medical information, is beyond human capability.
Yet, doctors and patients frequently make decisions by prioritizing crucial information and its significance. 
Consequently, the adoption of real-time speech summarization (RTSS) system is emerging as an effective approach to tackle this issue.

Compared to pre-recorded speech summarization, RTSS research has very little literature \cite{kameyama1996realtime_speech_sum}. 
Besides, in industry settings, to the best of our knowledge, there is currently no RTSS system deployed for real-world applications\footnote{In most papers, the term "real-time summarization" refers to the summarization of real-time news or events, instead of generating summaries in real-time.}.

In terms of medical domain, according to the latest survey by \cite{jain2022survey_medical_sum} and to the best of our knowledge, there is only one publicly available dataset for medical conversation summarization \cite{song2020medical_sum_dataset_chinese}.
This dataset consists of written text in the Chinese language and was crawled from an online healthcare service provider.
However, no speech summarization dataset for medical conversations is publicly available.

RTSS systems proposed by \cite{kameyama1996realtime_speech_sum} constantly update and revise the current summary state in the course of a dialogue using additional components, such as flexible recognizer of utterance units, utterance lookahead-er, and information overrider.
While these additions increase inference and training time, they also contribute to increased deployment and maintenance complexity.
Furthermore, from a business standpoint, these RTSS systems can degrade user experience as users are unaware of the exact moment when a comprehensive summary concludes.

To tackle all the problems above, we propose a new approach to a RTSS system for medical conversations.
Our contribution are as follows:
\begin{itemize}
    \item We propose the first deployable RTSS system for real-world applications.
    \item We introduce \textit{VietMed-Sum} - the first  speech summarization dataset for real-world medical conversations, to the best of our knowledge.
    \item We conduct the first attempt to leverage ChatGPT and human annotators colaboratively to create gold standard and synthetic summaries for medical conversations.
    \item We present baseline results on our dataset using various state-of-the-art models.
\end{itemize}

All code, data (English-translated and Vietnamese) and models are published online\footnote{\url{https://github.com/leduckhai/MultiMed/tree/master/VietMed-Sum}}%$^{,}$\footnote{\url{https://github.com/HySonLab/VietMed-Sum}}
.

\section{Real-time Speech Summarization System}

\begin{figure}
    \centering
    \includegraphics[width=0.45\textwidth]{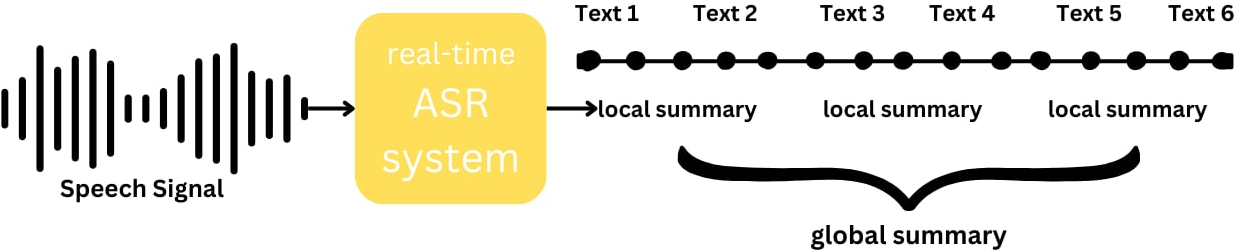}
    \caption{Visualization of our proposed RTSS}
    \label{fig:RTSS_diagram}
\end{figure}

\subsection{Previous Designs}
RTSS system proposed by \cite{kameyama1996realtime_speech_sum} has 3 major additional components: flexible recognizer of utterance units, utterance lookahead-er, and information overrider. 
Flexible recognizer of utterance units automatically split real-time Automatic Speech Recognition (ASR) transcript into segments with random lengths, while utterance lookahead-er seeks additional context from subsequent words generated by ASR, and information overrider continuously updates the summary in response to the latest contextual changes. We conducted surveys and gathered feedback from engineers and found that these extra components not only extend both inference and training time but also complicate RTSS systems, making it challenging for engineers to deploy and maintain them effectively.
Furthermore, the summary is constantly updated after each utterance generated by the ASR system.
% In other words, this RTSS system adds and updates the context, then generates a summary for every utterance recognized by the real-time ASR system.
This results in increased computational costs when compared to a scenario where a solid summary is generated after a set of utterances. The continuously updated summary creates confusion as users are unable to keep track, given the uncertainty of when a summary is completed.

\subsection{Our Design}
In contrast, our approach is much simpler.
Our design generates a local summary after every N utterances of speech within a conversation and a global summary after the end of a conversation.
Every local summary is generated using the corresponding local context of N utterances, without the need to continuously update the new context generated by real-time ASR utterances.
Meanwhile, the global summary serves as an "overrider" using the context of the entire conversation.

\subsection{Balance for System Delay}
RTSS system by \cite{kameyama1996realtime_speech_sum} generates a summary with a delay of one utterance.
A large number of delayed utterances results in longer waiting time for users to receive the generated summary. 
Conversely, a low number of delayed utterances means that the context necessary for accurate summaries is missing, making summarization unnecessary.
After analyzing the context within the \textit{VietMed} corpus \cite{vietmed_dataset} and conducting our internal user survey, we found that setting N = \{4, 5\} (or a maximum of around 30 seconds) strikes a suitable balance.
This ensures that each summary includes an adequate amount of context without keeping users waiting excessively.

\section{Data}
\label{section:data}

\subsection{Labeling strategy}
\label{subsection:labeling}
%\khai{DONE}

We used GPT-3.5 Turbo\footnote{https://platform.openai.com/docs/models/gpt-3-5-turbo} (or ChatGPT) to generate summaries for every transcript in our dataset, which we refer to as GPT-annotated summaries.
We then split the dataset into two subsets: \textbf{Gold standard (\textit{GOLD})} set and the \textbf{Synthetic  (\textit{SYN})} set. 
On the \textit{GOLD} set, we performed \textbf{human editing} where the human annotator edit the GPT summary according to the annotation guideline, while on the \textit{SYN} set, we did not. More information on GPT annotation is on section 4.

\subsection{Data Collection}
\label{subsection:data_collection}
%\khai{DONE}

\noindent \textbf{Real-world dataset (REAL)}: 
We choose the \textit{VietMed} dataset \cite{vietmed_dataset}, a real-world medical ASR dataset in Vietnamese, for annotating summaries.
This choice is driven by the fact that \textit{VietMed} currently stands as the world's largest and most generalizable publicly-available medical ASR dataset.

\noindent \textbf{Simulated dataset (SIM)}: To make the dataset more generalizable and to extend the scale of the existing \textit{VietMed} dataset, we used extra medical text data\footnote{https://github.com/duyvuleo/VNTC}.
We simulated real-world conversations by imitating the speaking style found in the \textit{VietMed} corpus. 
This includes incorporating hesitations, disfluencies, and stuttering words at a rate similar to that of \textit{VietMed} utterances.
% Also, we randomly trimmed both the beginning and the end of previously processed utterances to simulate the spoken text, and then resized them to match the average length of utterances in \textit{VietMed}.
Pseudo Python code for simulation is in the Appendix.

Our \textit{GOLD} set contains gold standard data from \textit{REAL} and \textit{SIM}, while \textit{SYN} contains GPT summaries from the extra medical text mentioned above.

% In this paper, we used GPT \cite{brown2020language} to generate summaries for a subset of SIM and the entire WEAK was used only for weakly-supervised training (not for recognition).

% We opted for human editing, not writing human summary from scratch, to reduce (1) annotator workload and (2) assist annotators in crafting high-quality summaries. We argue that by using GPT-generated summaries as a starting point, annotators can more efficiently navigate transcripts and focus on enhancing existing summaries, ensuring high-quality summaries overall. 

% the nuanced tone and stylistic preferences of the summaries should be kept consistently throughout the dataset. As such, . This approach to maintain the dataset consistency

\subsection{Annotation Process and Data Quality Control}
Details are in Data Annotation section in the Appendix.
% \khai{Is this part still necessary? Or maybe we should remove?}
% Following the process proposed by \cite{abacha2019MeQSum_dataset}, the third developer gave the following scores: 1 (perfect summary), 0.5 (acceptable), and 0 (incorrect, and replaced the summary in this case). 
% Based on these scores, the inter-annotator agreement was 95.2\%. 

% \subsection{Inter-annotator Agreement}
% \input{tables_and_figs/interannotator}

% To further control the quality, and to analyze inter-annotator agreement, we followed the process proposed by \cite{chen2021dialogsum_dataset}.
% For each dialogue in the test set we compared its three summaries: 1 by GPT, and 2 by the two developers, and computed their pair-wise ROUGE scores.
% Table \ref{interannotator} shows the inter-annotator agreement between GPT and two developers.

% (5) If the passage is broken or incomplete, and there is enough cluecomplete the missing ideas.

\subsection{Data Statistics}
%\khai{DONE}

\begin{table}[ht]
\centering
\resizebox{\columnwidth}{!}{%
\begin{tabular}{|l|ccc|c|c|}
\hline
\multirow{3}{*}{\textbf{}} & \multicolumn{3}{c|}{\textbf{Gold-standard Data}}                                                & \multirow{2}{*}{\textbf{Synthethic}} & \multirow{3}{*}{\textbf{All}} \\ \cline{2-4}
                           & \multicolumn{2}{c|}{\textbf{Real-world}}                                   & \textbf{Simulated} &                                      &                               \\ \cline{2-5}
                           & \multicolumn{1}{c|}{\textbf{Local}} & \multicolumn{1}{c|}{\textbf{Global}} & \textbf{Local}     & \textbf{Local}                       &                               \\ \hline
Train                      & \multicolumn{1}{c|}{837}            & \multicolumn{1}{c|}{382}             & 560                & 18981                                & 20760                         \\ \hline
Dev                        & \multicolumn{1}{c|}{874}            & \multicolumn{1}{c|}{624}             & 70                 & 0                                    & 1568                          \\ \hline
Test                       & \multicolumn{1}{c|}{1192}           & \multicolumn{1}{c|}{767}             & 70                 & 0                                    & 2029                          \\ \hline
Total                      & \multicolumn{1}{c|}{2903}           & \multicolumn{1}{c|}{1773}            & 700                & 18981                                & 24357                         \\ \hline
\end{tabular}
}
\caption{Data statistics for \textit{VietMed-Sum}. Full table is in the Appendix.}
\label{data_stats}
\end{table}
Table \ref{data_stats} shows the statistics of our dataset.
To construct our \textit{VietMed-Sum} dataset, we keep the original split of \textit{REAL} by \cite{vietmed_dataset} as 5-5-6 hours for the corresponding train-dev-test set.
We split our \textit{SIM} set with a ratio of 8:1:1 for the corresponding train-dev-test, while the entire \textit{SYN} is used for training. 

Acquisition and annotation of medical dataset is  challenging and costly, resulting in medical summarization datasets typically being smaller compared to those in the general domain.
Compared to other public medical written text summarization dataset, such as MeQSum corpus of summarized consumer health questions \cite{abacha2019MeQSum_dataset}, our dataset has 23 times more summaries.
Besides, compared to the Chinese medical text summarization dataset by \cite{song2020medical_sum_dataset_chinese}, ours is half the size.

\subsection{English-translated \textit{VietMed-Sum}}
We also introduce \textit{VietMed-Sum-en}, the English version of \textit{VietMed-Sum} which was translated using Google Translate\footnote{https://translate.google.com/}.
Results are in the Appendix.

\section{GPT for Annotation}
% As mentioned in \ref{section:data}, we used GPT to generate summaries for the dataset. In this section, we discuss experiments conducted to evaluate GPT's capabilities in generating Vietnamese medical summary.

\subsection{Motivation}
%\khai{DONE}

To the best of our knowledge, existing prominent Vietnamese summarization datasets, such as VietNews \cite{vietnews} and FAQSum \cite{minh2022vihealthbert}, utilize their titles and abstracts as summaries. 
Our dataset, however, lacks these pre-made summaries which would traditionally require human annotation. 
However, finding high-quality annotators for either low-resource languages like Vietnamese or medical domain is hard \cite{pavlick-etal-2014-language}.

In recent year, there has been an increasing focus on utilizing Large Language Models (LLMs) for annotation \cite{li2023coannotating,choi2024gpts,latif2023large}. Experimental results from \cite{DBLP:journals/corr/abs-2108-13487} showed that {fully GPT-3 labeling can ourperform fully human labeling in low-budget settings}. 
GPT has shown to have adequate medical knowledge \cite{meyer2024comparison}. 
Furthermore, it achieves reasonable performance as an annotator in sequence generation tasks in Vietnamese \cite{choi2024gpts}.

\subsection{GPT Setup}
We used GPT-3.5 Turbo to generate GPT summaries. 
Full setup details are in the Appendix.
% Our choice for the number of examples is motivated by balancing between the cost and the overall performance. Figure \ref{fig:prompt} illustrates the prompt we used.
% \khai{this could be thrown into the appendix, but more details of GPT hyperparams should help a lot for reproducibilty}

\subsection{Cost-Efficiency Evaluation}
\label{subsec:cost_eff_eval}

\begin{table}[!ht]
\centering
\resizebox{\columnwidth}{!}{%
\begin{tabular}{llccc}
\hline
\textbf{Cost} &\textbf{Method}& \textbf{R-1} & \textbf{R-2} & \textbf{R-L}\\
\hline
\multirow{2}{*}{$\$2.5$} &$\text{250 Human Summaries}$& 60.73& 45.35& 55.67\\

 &$\text{6k GPT Summaries}$& 56.69 & 40.13 & 50.61 \\
\hline
  &$\text{500 Human Summaries}$& 62.85& 47.19&57.50\\
%\multirow{3}{*}{$\$5$}  
{$\$5$}
&$\text{6k GPT}\rightarrow\text{250 Human-{reuse}}$& \textbf{62.88}& \textbf{47.81}&\textbf{57.67}\\
  % &$\text{GPT}\rightarrow\text{Human}$& \textbf{62.88}& \textbf{47.81}&\textbf{57.67}\\
&$\text{6k GPT}\rightarrow\text{250 Human-{new}}$& \textbf{63.45}& \textbf{47.65}&\textbf{57.49}\\

  \hline

\end{tabular}%
}
\caption{ROUGE on FAQSum on two budgets: \$2.5 and \$5. ViT5 is trained on the data from each method. $\text{GPT}\rightarrow\text{Human-{reuse}}$ refers to two-step finetuning on the 250 human summaries from the \$2.5-budget setting while $\text{GPT}\rightarrow\text{Human-{new}}$ refers to two-step finetuning on new 250 human summaries.}
\label{tab:vietmed_faq_performance}
\end{table}

\noindent Following the design from \cite{choi2024gpts, DBLP:journals/corr/abs-2108-13487}, we evaluated the performance difference of human-annotated summaries versus GPT-annotated summaries under a fixed budget. In particular, we evaluated ViT5 performance on FAQSum, a medical summarization dataset, trained with human-annotated summaries versus with GPT summaries on fixed budgets of \$2.5 and \$5, corresponding to 250 and 500 human-annotated summaries accordingly\footnote{Based on the minimum fee of \$0.01 per assignment on MTurk}. At \$2.5, GPT can generate around 6000 summaries\footnote{With an average of 700 input tokens and 20 output tokens per sample with rate \$0.50 per 1M input tokens and \$1.50 per 1M output tokens}.

The experimental results from the \$2.5-budget setting in Table \ref{tab:vietmed_faq_performance} demonstrate the importance of human annotation in the Vietnamese medical summarization task. Since summaries are heavily influenced by the annotators' medical knowledge and writing style, we hypothesize that GPT summaries still provide useful medical knowledge but require additional training on human summaries for writing style transfer. 

To verify our hypothesis, we devise the the \$5-budget setting where we fine-tuned ViT5 on GPT summaries, then on human summaries, and compared this two-step process with fine-tuning only on human summaries. We also doubled the number of human summaries for fair comparison. Experimental results show that the two-step process achieves slightly better results while costing significantly less time. We hypothesize that fine-tuning on GPT summaries helps provide medical knowledge and fine-tuning on human summaries helps the model aligns its output's text style.
\
\begin{table}[!ht]
\centering
\resizebox{\columnwidth}{!}{%
\begin{tabular}{lccc}
\hline
\textbf{Method}& \textbf{R-1} & \textbf{R-2} & \textbf{R-L}\\
\hline
$\text{6k Human Summaries}$& 65.80 & 50.82&60.83\\
{6k GPT Summaries}& 56.69 & 40.13 & 50.61 \\
%\multirow{3}{*}{$\$5$}  
$\text{6k GPT}\rightarrow\text{250 Human-{reuse}}$& \text{62.88}& \text{47.81}&\text{57.67}\\
$\text{6k GPT}\rightarrow\text{250 GPT}$& \text{56.93}& \text{39.81}&\text{50.14}\\
% $\text{6k GPT}\rightarrow\text{250 Human-{reuse}}$& \text{65.26}& \text{50.07}&\text{60.07}\\
  % &$\text{GPT}\rightarrow\text{Human}$& \text{62.88}& \text{47.81}&\text{57.67}\\
% $\text{6k GPT}\rightarrow\text{250 Human-{new}}$& \text{63.45}& \text{47.65}&\text{57.49}\\

  \hline

\end{tabular}%
}
\caption{ROUGE scores on FAQSum on 6k human summaries, 6k GPT summaries and the previously mentioned two-step finetuning process of ViT5.}
\label{tab:faq_6k}
\end{table}

Table \ref{tab:faq_6k} further support our argument. When we perform two-step finetuning on ViT5, the performance gap between human summaries and GPT summaries is significantly closed. As such, our labeling strategy involves creating the \textit{SYN} set and refining the \textit{GOLD} set to balance between time-consumption, cost and performance.

\subsection{Human Evaluation}
We further investigate the characteristics of the GPT summaries empirically. In this experiment, we sample 50 arbitrary transcripts of varying lengths from \textit{VietMed-Sum} and let two annotators independently summarize them, one with GPT summaries as references (human editing) and one without.  

\noindent \textbf{Human Editing Time-Efficiency}:
We quantify the overall improvement in time taken to annotate the transcripts between human editing and manually writing the summary. We found that annotators who perform human editing is approximately 70\% faster than those who perform manual summary writing. As such, we perform human editing on the \textit{GOLD} set.

\noindent \textbf{Hallucination}:
To make sure the GPT summary is factually aligned with the transcript, we further cross-check the GPT summaries with the transcripts.
We found that (1) the GPT summaries sometimes contain details implied but not explicitly mentioned in the transcript and (2) GPT is easily confused by transcripts with a lot of spoken language characteristics (i.e. hesitations, disfluencies and stuttering words). This typically leads to hallucinate and we found that around 25\% of our samples have hallucination. Furthermore, the GPT summaries are often more lengthy with an average compression rate of 30\%, higher than required in the guideline. As such, we ask the annotators to strictly adhere to the annotation guideline when editing.

\section{Experimental Setup}

% In particular, each model is fine-tuned using AdamW with learning rate 1e-5 and batch size 16 over 30 epochs using early stopping with a patience of 3. \khai{hyperparams could be thrown into appendix, but needs more details here, all hyperparams should be reported}
% \khai{remember to mention about 3 seeds per training!}
\subsection{Evaluation Metrics}
We use ROUGE \cite{lin2004rouge}, a metric commonly used for summarization, to evaluate our models. More details are in the Appendix.

\subsection{Baseline Summarization Models}
We employed $\text{BARTpho}_{\text{syllable}}$ and  $\text{BARTpho}_{\text{word}}$ \cite{tran2022bartpho}, ViT5, ViT5-vietnews \cite{phan2022vit5} (ViT5 fine-tuned on Vietnews summarization dataset \cite{8573420}), and ViPubmedT5 \cite{phan2023enriching} models in our experiments. 
More information about the models is in the Appendix.

\subsection{Downstream Tasks}
\noindent \textbf{Summarization on Human Transcript}:
We train the models on the abstractive summarization task on VietMed-Sum. 
To evaluate their performance, we calculate their ROUGE scores on the local and global summaries in the test set.

\noindent \textbf{Summarization on ASR Transcript}:
We also evaluated the models' performance from transcripts obtained from audio speech recognition (ASR). 
We employed the best ASR model on \textit{VietMed} from \cite{vietmed_dataset} with a Word-Error-Rate (WER) of 28.8\% to generate the ASR transcripts for summarization. 
This creates noisier text which is more challenging for the baseline models.

\section{Experimental Results}
 We report the ROUGE of our baseline models on the Global summaries and Local summaries subset from \textit{GOLD}.
\subsection{Gold Standard Data Summarization}
%\khai{DONE}

% \begin{table}[ht]
% \centering
% \begin{tabular}{llccc}
% \hline
% Model & R-1 & R-2 & R-L \\ 
% \hline
% $\text{BARTpho}_{\text{syllable}}$ & 58.94 & 36.92 & 46.53 \\
% $\text{BARTpho}_{\text{word}}$ & 57.43 & 36.09 & 45.73 \\
% ViPubmedT5 & 59.91 & 37.3 & 46.7 \\
% ViT5 & 59.79 & 37.48 & \textbf{47.47} \\
% ViT5-vietnews & \textbf{59.97 }& \textbf{37.66} & 47.2 \\
% \hline
% \end{tabular}
% \caption{Model performance on Train Whole dataset.}
% \label{tab:performance_train_whole}
% \end{table}
\begin{table}[ht]
\centering
\resizebox{\columnwidth}{!}{%
\begin{tabular}{l|ccc|ccc}
\toprule
\textbf{Data} & \multicolumn{3}{c|}{\textbf{Global Summaries}} & \multicolumn{3}{c}{\textbf{Local Summaries}} \\
% \cmidrule{2-7}
\textbf{GOLD} & \textbf{R-1} & \textbf{R-2} & \textbf{R-L} & \textbf{R-1} & \textbf{R-2} & \textbf{R-L} \\
\midrule
$\text{BARTpho}_{\text{syllable}}$ & 60.92 & 40.71 & 49.38 & 59.07 & 38.27 & 47.69 \\
$\text{BARTpho}_{\text{word}}$ & 58.83 & 39.71 & 48.07 & 57.76 & 37.43 & 46.87 \\
ViT5 & 61.65 & 40.56 & 49.62 & 59.95 & 38.66 & 48.66 \\
ViT5-vietnews & 61.90 & 41.07 & 49.61 & 59.94 & 38.69 & 48.25 \\
ViPubmedT5 & 61.73 & 40.17 & 48.81 & 59.99 & 38.30 & 47.67 \\
\bottomrule
\end{tabular}%
}
\caption{Experimental results on VietMed-Sum's \textit{GOLD} test set of each model fine-tuned on local + global summaries of \textit{GOLD}.}
\label{tab:performance_train_whole}
\end{table}

Table \ref{tab:performance_train_whole} shows the ROUGE scores on our \textit{GOLD} test set for the baseline models fine-tuned on the combination of \textit{GOLD} local and global summaries. 
ViT5, ViPubmedT5, and ViT5-vietnews consistently outperforms the BARTpho variants. 
% Although ViPubmedT5 was pre-trained on a very large-scale biomedical text, its performance is similar to that of ViT5 and ViT5-vietnews.

% \subsection{Ablation Studies}

% \begin{table}[ht]
% \centering
% \begin{tabular}{llccc}
% \hline
%  Data & Model & R-1 & R-2 & R-L \\ 
% \hline
% Global Summaries &$\text{BARTpho}_{\text{syllable}}$ & \textbf{58.77} & \textbf{36.4} & \textbf{46.26} \\
% &$\text{BARTpho}_{\text{word}}$ & 57.31 & 34.96 & 45.07 \\
% &ViPubmedT5 & 56.68 & 33.85 & 43.72 \\
% &ViT5 & 56.85 & 33.97 & 44.34 \\
% &ViT5-vietnews & 58.58 & 35.7 & 45.54 \\
% \hline
% Local Summaries-segment & $\text{BARTpho}_{\text{syllable}}$ & 57.14 & 36.12 & 45.16 \\
% & $\text{BARTpho}_{\text{word}}$ & 55.09 & 34.88 & 43.85 \\
% & ViPubmedT5 & 58.26 & 36.12 & 45.9 \\
% & ViT5 & 58.43 & 37.04 & 46.55 \\
% & ViT5-vietnews & \textbf{59.49} & \textbf{37.98} & \textbf{47.21 }\\
% \hline

% \end{tabular}
% \caption{Model performance on Train Global Summaries dataset.}
% \label{tab:performance_train_full_dialogue}
% \end{table}

\begin{table}[ht]
\centering
\resizebox{\columnwidth}{!}{%
\begin{tabular}{l|ccc|ccc}
\toprule
\textbf{Data} & \multicolumn{3}{c|}{\textbf{Global Summaries}} & \multicolumn{3}{c}{\textbf{Local Summaries}} \\
% \midrule

\textbf{Global} & \textbf{R-1} & \textbf{R-2} & \textbf{R-L} & \textbf{R-1} & \textbf{R-2} & \textbf{R-L} \\
\hline
$\text{BARTpho}_{\text{syllable}}$ & 61.69 & 41.19 & 49.87 & 59.13 & 37.80 & 47.59 \\
$\text{BARTpho}_{\text{word}}$ & 59.45 & 39.05 & 47.79 & 57.8 & 36.28 & 46.35 \\
ViT5 & 58.89 & 37.2 & 46.76 & 56.79 & 34.97 & 45.31 \\
ViT5-vietnews & 60.67 & 39.20 & 48.17 & 58.64 & 36.83 & 46.61 \\
ViPubmedT5 & 58.57 & 36.65 & 45.75 & 56.83 & 34.70 & 44.72 \\
\midrule
\textbf{Local} & \textbf{R-1} & \textbf{R-2} & \textbf{R-L} & \textbf{R-1} & \textbf{R-2} & \textbf{R-L} \\
\midrule
$\text{BARTpho}_{\text{syllable}}$ & 58.33 & 39.62 & 47.49 & 57.37 & 37.43 & 46.41 \\
$\text{BARTpho}_{\text{word}}$ & 56.46 & 38.62 & 46.20 & 55.65 & 36.36 & 45.13 \\
ViT5 & 59.82 & 40.47 & 48.74 & 59.11 & 38.41 & 47.87 \\
ViT5-vietnews & 61.04 & 41.43 & 49.57 & 60.10 & 39.39 & 48.52 \\
ViPubmedT5 & 59.00 & 38.27 & 47.36 & 58.99 & 37.38 & 47.11 \\
\bottomrule
\end{tabular}%
}
\caption{Results on VietMed-Sum's Global and Local subset of \textit{GOLD} test set. Each baseline model is fine-tuned on the global summaries (above) and the local summaries (below) of \textit{GOLD}}
\label{tab:performance_train_full_dialogue}
\end{table}

Results from Table \ref{tab:performance_train_full_dialogue} shows the models have a noticeable drop in performance. 
On the local summaries, the ROUGE scores from the BARTpho variants are much lower than that of the ViT5 variants. 
Conversely, when fine-tuned on the global summaries, both variants of BARTpho performs much better than the ViT5 variants except ViT5-vietnews, probably because ViT5-vietnews was previously fine-tuned on other Vietnamese abstractive summarization dataset.

\subsection{Synthetic Data Summarization}
\begin{table}[!ht]
\centering
\resizebox{\columnwidth}{!}{%
\begin{tabular}{l|ccc|ccc}
\toprule
\textbf{Data} & \multicolumn{3}{c|}{\textbf{Global Summaries}} & \multicolumn{3}{c}{\textbf{Local Summaries}} \\
% \midrule
\textbf{SYN} & \textbf{R-1} & \textbf{R-2} & \textbf{R-L} & \textbf{R-1} & \textbf{R-2} & \textbf{R-L} \\
\midrule
$\text{BARTpho}_{\text{syllable}}$ & 60.37 & 40.27 & 48.48 & 58.68 & 38.17 & 47.14 \\
$\text{BARTpho}_{\text{word}}$ & 59.64 & 40.21 & 48.01 & 57.50 & 37.31 & 46.01 \\
ViT5 & 61.71 & 42.36 & 50.17 & 59.74 & 39.83 & 48.65 \\
ViT5-vietnews & 60.78 & 40.79 & 49.07 & 58.18 & 38.21 & 47.01 \\
ViPubmedT5 & 60.58 & 40.95 & 48.72 & 58.44 & 38.40 & 47.09 \\
\midrule
\textbf{SYN + GOLD} & \textbf{R-1} & \textbf{R-2} & \textbf{R-L} & \textbf{R-1} & \textbf{R-2} & \textbf{R-L} \\
\midrule
$\text{BARTpho}_{\text{syllable}}$ & 61.13 & 41.55 & 49.83 & 59.10 & 39.10 & 48.06 \\
$\text{BARTpho}_{\text{word}}$ & 61.11 & 42.08 & 49.66 & 59.16 & 39.36 & 48.12 \\
ViT5 & 63.23 & 43.92 & 51.64 & 60.9 & 40.93 & 49.83 \\
ViT5-vietnews & 62.68 & 43.59 & 51.58 & 60.31 & 40.52 & 49.32 \\
ViPubmedT5 & 62.15 & 42.92 & 50.45 & 59.96 & 40.22 & 48.82 \\
\midrule
\textbf{SYN $\rightarrow$ GOLD} & \textbf{R-1} & \textbf{R-2} & \textbf{R-L} & \textbf{R-1} & \textbf{R-2} & \textbf{R-L} \\
\midrule
$\text{BARTpho}_{\text{syllable}}$ & 62.37 & 41.87 & 50.49 & 60.5 & 39.78 & 49.24 \\
$\text{BARTpho}_{\text{word}}$ & 60.93 & 41.80 & 49.89 & 59.38 & 39.43 & 48.67 \\
\textbf{ViT5} & \textbf{64.52} & \textbf{45.12} & \textbf{52.95} & \textbf{62.56} & \textbf{42.41} & \textbf{51.61} \\
\textbf{ViT5-vietnews} & \textbf{63.34} & \textbf{43.29} & \textbf{51.58} & \textbf{61.73} & \textbf{41.20} & \textbf{50.23} \\
ViPubmedT5 & 61.70 & 40.13 & 48.80 & 59.99 & 38.31 & 47.68 \\
\bottomrule
\end{tabular}%
}
\caption{Experimental results on VietMed-Sum's \textit{GOLD} test set of each baseline model fine-tuned on \textit{SYN}, \textit{(SYN + GOLD)} and \textit{SYN $\rightarrow$ GOLD}. + refers to concatenating the datasets while $\rightarrow$ refers to two-step fine-tuning. Bolded text refers to two best performing models.}
\label{tab:comparison_data5_data6}
\end{table}

Table \ref{tab:comparison_data5_data6} shows the ROUGE scores of each model fine-tuned on \textit{SYN}-only and with \textit{GOLD}. 
We found that fine-tuning only on \textit{SYN} data did not improve the models' performance. 
However, when we incorporated \textit{GOLD} into fine-tuning the models either by concatenating the \textit{GOLD} and \textit{SYN} data or by performing two-step fine-tuning \textit{SYN} $\rightarrow$ \textit{GOLD}, the performance of the models drastically improved compared to training only on \textit{GOLD} on all metrics. 
This is consistent with our observations from Subsection \ref{subsec:cost_eff_eval}. 
% Finally, we observe that ViT5 outperformed ViT5-vietnews in both settings.

% Nonetheless, this highlights the potential of distilling LLMs capabilities to smaller models in this task. Finally, we observe that ViT5 outperforms ViT5-vietnews in both settings, probably because the pre-trained ViT5 with enough training samples converges faster than the previously fine-tuned ViT5-vietnews.

\subsection{ASR Transcript Summarization}
\begin{table}[!ht]
\centering

\resizebox{\columnwidth}{!}{%
% \begin{tabular}{llccc}
% \toprule
% \textbf{Model} & \textbf{R-1} & \textbf{R-2} & \textbf{R-L} \\
\begin{tabular}{l|ccc|ccc}
\toprule
\multirow{2}{*}{\textbf{Model}} & \multicolumn{3}{c|}{\textbf{Global Summaries}} & \multicolumn{3}{c}{\textbf{Local Summaries}} \\

% \midrule
% whole & bartpho-syllable & 55.32 & 32.46 & 44.04 \\
%  & bartpho-word & 53.22 & 30.98 & 42.12 \\
%  & ViT5 & 54.86 & 31.70 & 43.76 \\
%  & ViT5-vietnews & 55.33 & 31.93 & 43.68 \\
%  & vipubmedt5 & 55.64 & 31.80 & 43.42 \\
% \midrule
% Local Summaries\_segment & bartpho-syllable & 54.44 & 31.41 & 43.16 \\
%  & bartpho-word & 52.96 & 29.69 & 41.49 \\
%  & ViT5 & 52.81 & 29.49 & 41.29 \\
%  & ViT5-vietnews & 54.34 & 31.03 & 42.61 \\
%  & vipubmedt5 & 52.34 & 28.09 & 39.97 \\
% \midrule
% Global Summaries & bartpho-syllable & 53.69 & 31.75 & 43.01 \\
%  & bartpho-word & 51.62 & 30.11 & 40.64 \\
%  & ViT5 & 55.09 & 32.55 & 43.88 \\
%  & ViT5-vietnews & 54.72 & 32.28 & 43.37 \\
%  & vipubmedt5 & 53.95 & 31.33 & 42.42 \\
% \midrule
% 19k & bartpho-syllable & 54.31 & 31.17 & 42.23 \\
%  & bartpho-word & 53.76 & 31.18 & 41.92 \\
%  & ViT5 & 55.21 & 32.47 & 43.29 \\
%  & ViT5-vietnews & 54.01 & 31.49 & 41.85 \\
%  & vipubmedt5 & 54.28 & 31.00 & 42.41 \\
% \midrule
% 20k & bartpho-syllable & 54.78 & 31.92 & 43.46 \\
%  & bartpho-word & 54.06 & 31.58 & 42.56 \\
%  & ViT5 & 56.52 & 34.30 & 45.15 \\
%  & ViT5-vietnews & 58.21 & 35.67 & 45.84 \\
%  & vipubmedt5 & 58.30 & 36.07 & 45.91 \\
% \midrule
% 19k $\rightarrow$ whole & bartpho-syllable & 55.46 & 32.57 & 44.38 \\
%  & bartpho-word & 54.22 & 32.07 & 43.27 \\
% \textbf{Model} & \textbf{R-1} & \textbf{R-2} & \textbf{R-L} \\
 & \textbf{R-1} & \textbf{R-2} & \textbf{R-L} & \textbf{R-1} & \textbf{R-2} & \textbf{R-L} \\
\midrule
ViT5 & 58.95 & 36.82 & 46.63 & 56.78 & 34.43 & 45.52 \\
ViT5-vietnews & 58.22 & 35.34 & 46.02 & 56.48 & 33.65 & 45.10 \\
 % & vipubmedt5-base & 55.48 & 31.57 & 43.06 \\
\bottomrule
\end{tabular}
}

\caption{Experimental results on ASR transcripts of the two best performing models from Section 6.2.
Full table with all results is in the Appendix.}
\label{tab:asr_result}
\end{table}

We report the results of our baseline models on the ASR transcripts on Table \ref{tab:asr_result}. The performance of the models is worse than that on \textit{VietMed-Sum}, which we attribute to the noisy nature of the text generated by the ASR models. Nevertheless, the ROUGE scores remain fairly reasonable, which is proof to our model's robustness.

% \subsection{VietMed-Sum-en}
% % \input{tables_and_figs/en}
% We also introduce VietMed-Sum-en, the English version of VietMed-Sum which was translated using Google Translate.
% in Table \ref{tab:comparison_en}.
\subsection{Human Evaluation}
\begin{table}
\centering
\resizebox{\columnwidth}{!}{%
\begin{tabular}{|l|c|c|c|c|}
\hline
\textbf{Summary} & \textbf{Fluency} & \textbf{Consistency} & \textbf{Relevance} & \textbf{Coherance} \\ \hline
ChatGPT               & 3.8              & 3.3                  & 5.0                & 4.2                \\ \hline
GOLD                  & 5.0              & 5.0                  & 5.0                & 5.0                \\ \hline
ViT5                   & 4.0              & 4.2                  & 4.3                & 4.2                \\ \hline
\end{tabular}%
}
\caption{Results for human evaluation on 50 samples. Scores range from 1 (worst) to 5 (best). \textit{GOLD} is the baseline which has all scores of 5. ViT is the best model for ROUGE scores.}
\label{human_eval}
\end{table}
While ROUGE is commonly used to evaluate the performance of the models, does not measure the fluency and factual alignment of the summaries. 
As such, we adopt the human evaluation methodology from \cite{chen2021dialogsum_dataset, kryscinski2019human_eval_metric_sum} and report the results in Table \ref{human_eval}.
Details of experiments are in the Appendix.

\section{Conclusion}
In this work, we propose a novel RTSS system that generates a local summary after every N utterances within a conversation and a global summary for the entire conversation. 
Unlike previous works that continuously update the summary after each utterance generated by ASR systems which might be hard for users to follow, our system could improve user experience and lower computational costs.
Secondly, we present \textit{VietMed-Sum}, the first speech summarization dataset for medical conversations.  
Thirdly, our proposed labeling strategy strikes a balance between performance of summarization models, annotation cost, and annotation time (approximately 70\% time reduction).
We report the use of our proposed synthetic data generated by LLM, which improves models' performance across all metrics.
Notable is an average improvement of 2.74 in the R-1 score for ViT5.

\section{Acknowledgement}
We thank Bao Tran at Chubb Canada and Linh Nguyen from Bucknell University for helping the initial annotation.
We appreciate David Thulke at RWTH Aachen University and AppTek GmbH for his precious feedback.
We thank Ralf Schlüter at RWTH Aachen University for supporting computing resource to conduct experiments.

\bibliographystyle{IEEEtran}
\bibliography{mainbib}

\begin{thebibliography}{10}
\providecommand{\url}[1]{#1}
\csname url@samestyle\endcsname
\providecommand{\newblock}{\relax}
\providecommand{\bibinfo}[2]{#2}
\providecommand{\BIBentrySTDinterwordspacing}{\spaceskip=0pt\relax}
\providecommand{\BIBentryALTinterwordstretchfactor}{4}
\providecommand{\BIBentryALTinterwordspacing}{\spaceskip=\fontdimen2\font plus
\BIBentryALTinterwordstretchfactor\fontdimen3\font minus \fontdimen4\font\relax}
\providecommand{\BIBforeignlanguage}[2]{{%
\expandafter\ifx\csname l@#1\endcsname\relax
\typeout{** WARNING: IEEEtran.bst: No hyphenation pattern has been}%
\typeout{** loaded for the language `#1'. Using the pattern for}%
\typeout{** the default language instead.}%
\else
\language=\csname l@#1\endcsname
\fi
#2}}
\providecommand{\BIBdecl}{\relax}
\BIBdecl

\bibitem{kameyama1996realtime_speech_sum}
M.~Kameyama, G.~Kawai, and I.~Arima, ``A real-time system for summarizing human-human spontaneous spoken dialogues,'' in \emph{Proceeding of Fourth International Conference on Spoken Language Processing. ICSLP'96}, 1996.

\bibitem{jain2022survey_medical_sum}
R.~Jain, A.~Jangra, S.~Saha, and A.~Jatowt, ``A survey on medical document summarization,'' \emph{arXiv preprint arXiv:2212.01669}, 2022.

\bibitem{song2020medical_sum_dataset_chinese}
Y.~Song, Y.~Tian, N.~Wang, and F.~Xia, ``Summarizing medical conversations via identifying important utterances,'' in \emph{Proceedings of the 28th International Conference on Computational Linguistics}, 2020, pp. 717--729.

\bibitem{vietmed_dataset}
K.~Le-Duc, ``Vietmed: A dataset and benchmark for automatic speech recognition of vietnamese in the medical domain,'' in \emph{Proceedings of the 2024 Joint International Conference on Computational Linguistics, Language Resources and Evaluation (LREC-COLING 2024)}, 2024, pp. 17\,365--17\,370.

\bibitem{abacha2019MeQSum_dataset}
A.~B. Abacha and D.~Demner-Fushman, ``On the summarization of consumer health questions,'' in \emph{Proceedings of the 57th Annual Meeting of the Association for Computational Linguistics}, 2019, pp. 2228--2234.

\bibitem{vietnews}
V.-H. Nguyen, T.-C. Nguyen, M.-T. Nguyen, and N.~X. Hoai, ``Vnds: A vietnamese dataset for summarization,'' in \emph{2019 6th NAFOSTED Conference on Information and Computer Science (NICS)}, 2019, pp. 375--380.

\bibitem{minh2022vihealthbert}
\BIBentryALTinterwordspacing
N.~Minh, V.~H. Tran, V.~Hoang, H.~D. Ta, T.~H. Bui, and S.~Q.~H. Truong, ``{V}i{H}ealth{BERT}: Pre-trained language models for {V}ietnamese in health text mining,'' in \emph{Proceedings of the Thirteenth Language Resources and Evaluation Conference}, N.~Calzolari, F.~B{\'e}chet, P.~Blache, K.~Choukri, C.~Cieri, T.~Declerck, S.~Goggi, H.~Isahara, B.~Maegaard, J.~Mariani, H.~Mazo, J.~Odijk, and S.~Piperidis, Eds.\hskip 1em plus 0.5em minus 0.4em\relax Marseille, France: European Language Resources Association, Jun. 2022, pp. 328--337. [Online]. Available: \url{https://aclanthology.org/2022.lrec-1.35}
\BIBentrySTDinterwordspacing

\bibitem{pavlick-etal-2014-language}
\BIBentryALTinterwordspacing
E.~Pavlick, M.~Post, A.~Irvine, D.~Kachaev, and C.~Callison-Burch, ``The language demographics of {A}mazon {M}echanical {T}urk,'' \emph{Transactions of the Association for Computational Linguistics}, vol.~2, pp. 79--92, 2014. [Online]. Available: \url{https://aclanthology.org/Q14-1007}
\BIBentrySTDinterwordspacing

\bibitem{li2023coannotating}
M.~Li, T.~Shi, C.~Ziems, M.-Y. Kan, N.~F. Chen, Z.~Liu, and D.~Yang, ``Coannotating: Uncertainty-guided work allocation between human and large language models for data annotation,'' 2023.

\bibitem{choi2024gpts}
J.~Choi, E.~Lee, K.~Jin, and Y.~Kim, ``Gpts are multilingual annotators for sequence generation tasks,'' 2024.

\bibitem{latif2023large}
S.~Latif, M.~Usama, M.~I. Malik, and B.~W. Schuller, ``Can large language models aid in annotating speech emotional data? uncovering new frontiers,'' 2023.

\bibitem{DBLP:journals/corr/abs-2108-13487}
\BIBentryALTinterwordspacing
S.~Wang, Y.~Liu, Y.~Xu, C.~Zhu, and M.~Zeng, ``Want to reduce labeling cost? {GPT-3} can help,'' \emph{CoRR}, vol. abs/2108.13487, 2021. [Online]. Available: \url{https://arxiv.org/abs/2108.13487}
\BIBentrySTDinterwordspacing

\bibitem{meyer2024comparison}
A.~Meyer, J.~Riese, and T.~Streichert, ``Comparison of the performance of gpt-3.5 and gpt-4 with that of medical students on the written german medical licensing examination: Observational study,'' \emph{JMIR Medical Education}, vol.~10, p. e50965, 2024.

\bibitem{lin2004rouge}
C.-Y. Lin, ``Rouge: A package for automatic evaluation of summaries,'' in \emph{Text summarization branches out}, 2004, pp. 74--81.

\bibitem{tran2022bartpho}
N.~L. Tran, D.~M. Le, and D.~Q. Nguyen, ``Bartpho: Pre-trained sequence-to-sequence models for vietnamese,'' 2022.

\bibitem{phan2022vit5}
L.~Phan, H.~Tran, H.~Nguyen, and T.~H. Trinh, ``Vit5: Pretrained text-to-text transformer for vietnamese language generation,'' 2022.

\bibitem{8573420}
M.-T. Nguyen, H.-D. Nguyen, T.-H.-N. Nguyen, and V.-H. Nguyen, ``Towards state-of-the-art baselines for vietnamese multi-document summarization,'' in \emph{2018 10th International Conference on Knowledge and Systems Engineering (KSE)}, 2018, pp. 85--90.

\bibitem{phan2023enriching}
L.~Phan, T.~Dang, H.~Tran, T.~H. Trinh, V.~Phan, L.~D. Chau, and M.-T. Luong, ``Enriching biomedical knowledge for low-resource language through large-scale translation,'' in \emph{Proceedings of the 17th Conference of the European Chapter of the Association for Computational Linguistics}, 2023, pp. 3131--3142.

\bibitem{chen2021dialogsum_dataset}
Y.~Chen, Y.~Liu, L.~Chen, and Y.~Zhang, ``Dialogsum: A real-life scenario dialogue summarization dataset,'' in \emph{Findings of the Association for Computational Linguistics: ACL-IJCNLP 2021}, 2021, pp. 5062--5074.

\bibitem{kryscinski2019human_eval_metric_sum}
W.~Kryscinski, N.~S. Keskar, B.~McCann, C.~Xiong, and R.~Socher, ``Neural text summarization: A critical evaluation,'' in \emph{Proceedings of the 2019 Conference on Empirical Methods in Natural Language Processing and the 9th International Joint Conference on Natural Language Processing (EMNLP-IJCNLP)}.\hskip 1em plus 0.5em minus 0.4em\relax Association for Computational Linguistics, 2019.

\bibitem{wolf2019huggingface}
T.~Wolf, L.~Debut, V.~Sanh, J.~Chaumond, C.~Delangue, A.~Moi, P.~Cistac, T.~Rault, R.~Louf, M.~Funtowicz \emph{et~al.}, ``Huggingface's transformers: State-of-the-art natural language processing,'' \emph{arXiv preprint arXiv:1910.03771}, 2019.

\bibitem{lewis2019bart}
M.~Lewis, Y.~Liu, N.~Goyal, M.~Ghazvininejad, A.~Mohamed, O.~Levy, V.~Stoyanov, and L.~Zettlemoyer, ``Bart: Denoising sequence-to-sequence pre-training for natural language generation, translation, and comprehension,'' 2019.

\bibitem{raffel2020exploring}
C.~Raffel, N.~Shazeer, A.~Roberts, K.~Lee, S.~Narang, M.~Matena, Y.~Zhou, W.~Li, and P.~J. Liu, ``Exploring the limits of transfer learning with a unified text-to-text transformer,'' \emph{The Journal of Machine Learning Research}, vol.~21, no.~1, pp. 5485--5551, 2020.

\bibitem{conneau2019unsupervised}
A.~Conneau, K.~Khandelwal, N.~Goyal, V.~Chaudhary, G.~Wenzek, F.~Guzm{\'a}n, E.~Grave, M.~Ott, L.~Zettlemoyer, and V.~Stoyanov, ``Unsupervised cross-lingual representation learning at scale,'' \emph{arXiv preprint arXiv:1911.02116}, 2019.

\bibitem{roberts2022t5x}
A.~Roberts, H.~W. Chung, A.~Levskaya, G.~Mishra, J.~Bradbury, D.~Andor, S.~Narang, B.~Lester, C.~Gaffney, A.~Mohiuddin, C.~Hawthorne, A.~Lewkowycz, A.~Salcianu, M.~van Zee, J.~Austin, S.~Goodman, L.~B. Soares, H.~Hu, S.~Tsvyashchenko, A.~Chowdhery, J.~Bastings, J.~Bulian, X.~Garcia, J.~Ni, A.~Chen, K.~Kenealy, J.~H. Clark, S.~Lee, D.~Garrette, J.~Lee-Thorp, C.~Raffel, N.~Shazeer, M.~Ritter, M.~Bosma, A.~Passos, J.~Maitin-Shepard, N.~Fiedel, M.~Omernick, B.~Saeta, R.~Sepassi, A.~Spiridonov, J.~Newlan, and A.~Gesmundo, ``Scaling up models and data with $\texttt{t5x}$ and $\texttt{seqio}$,'' 2022.

\bibitem{flax2020github}
\BIBentryALTinterwordspacing
J.~Heek, A.~Levskaya, A.~Oliver, M.~Ritter, B.~Rondepierre, A.~Steiner, and M.~van {Z}ee, ``{F}lax: A neural network library and ecosystem for {JAX},'' 2023. [Online]. Available: \url{http://github.com/google/flax}
\BIBentrySTDinterwordspacing

\end{thebibliography}

% -----------Supplementary Material-----------
\clearpage % Page break for arxiv version
\onecolumn
\appendix

\section{Limitations}
We acknowledge that our work relies on GPT which may contain hidden and unknown biases. As GPT is regularly updated, parts of our experiments may not be exactly emulated. 

\section{Additional Details about Experiments}

\subsection{Additional Details of Data Collection: Data Simulation Methods}
Pseudo Python code for creating simulated data (\textit{SIM}) using the extra medical text data, which is described in the subsection Data Collection \ref{subsection:data_collection}.

\begin{python}
fillers = [List of Vietnamese filler words]

def simulate_speaking_style(words, fillers):
   new_words = []
   
   for word in words:
      # Randomly repeat a word with 0.01 probability
      if random.random() < 0.01:
         new_words.append(word)
      # Randomly insert a filler with 0.01 probability
      if random.random() < 0.01:
         new_words.append(random.choice(fillers))
      
   return ' '.join(new_words)   
\end{python}

\begin{python}
def simulate_spoken_text(simulated_speaking_style_utterances, avg_lengths):
   
   chosen_len = random.choice(avg_lengths)
   
   # Randomly decide trimming strategy
   trim_strategy = random.choice(['back', 'front', 'both'])
      
   if trim_strategy == 'back':
      trimmed_words = words[:chosen_len]
   elif trim_strategy == 'front':
      trimmed_words = words[-chosen_len:]
   elif trim_strategy == 'both':
      trimmed_words = words[flexible_start:flexible_end]

   return trimmed_words
\end{python}

\subsection{Additional Data Statistics}
Table \ref{appx_data_stats} shows the additional data statistics which extends the Table \ref{data_stats}.
\begin{table}[ht]
\centering
%\resizebox{\columnwidth}{!}{%
\begin{tabular}{lcllcl}
\cline{1-5}
\multicolumn{1}{|l|}{}                                    & \multicolumn{3}{c|}{\textbf{Gold-standard Data}}                                                                     & \multicolumn{1}{c|}{}                                      &                                               \\ \cline{2-4}
\multicolumn{1}{|l|}{}                                    & \multicolumn{2}{c|}{\textbf{Real-world}}                                   & \multicolumn{1}{c|}{\textbf{Simulated}} & \multicolumn{1}{c|}{\multirow{-2}{*}{\textbf{Synthethic}}} &                                               \\ \cline{2-5}
\multicolumn{1}{|l|}{\multirow{-3}{*}{}} & \multicolumn{1}{c|}{\textbf{Local}} & \multicolumn{1}{c|}{\textbf{Global}} & \multicolumn{1}{c|}{\textbf{Local}}     & \multicolumn{1}{c|}{\textbf{Local}}                        &                                               \\ \cline{1-5}
\multicolumn{1}{|l|}{Train}                               & \multicolumn{1}{c|}{837}            & \multicolumn{1}{c|}{382}             & \multicolumn{1}{c|}{560}                & \multicolumn{1}{c|}{18981}                                 &                                               \\ \cline{1-5}
\multicolumn{1}{|l|}{Dev}                                 & \multicolumn{1}{c|}{874}            & \multicolumn{1}{c|}{624}             & \multicolumn{1}{c|}{70}                 & \multicolumn{1}{c|}{0}                                     &                                               \\ \cline{1-5}
\multicolumn{1}{|l|}{Test}                                & \multicolumn{1}{c|}{1192}           & \multicolumn{1}{c|}{767}             & \multicolumn{1}{c|}{70}                 & \multicolumn{1}{c|}{0}                                     &                                               \\ \hline
\multicolumn{1}{|l|}{Total}                               & \multicolumn{1}{c|}{2903}           & \multicolumn{1}{c|}{1773}            & \multicolumn{1}{c|}{700}                & \multicolumn{1}{c|}{18981}                                 & \multicolumn{1}{c|}{24357}                    \\ \hline
\multicolumn{6}{c}{\textbf{Appendix}}                                                                                                                                                                                                                                 \\ \hline
\multicolumn{1}{|l|}{\#Summary words}                & \multicolumn{1}{c|}{46683}          & \multicolumn{1}{c|}{40465}           & \multicolumn{1}{c|}{18235}              & \multicolumn{1}{c|}{669900}                                & \multicolumn{1}{c|}{775283}                   \\ \hline
\multicolumn{1}{|l|}{\#Input words}               & \multicolumn{1}{c|}{202944}         & \multicolumn{1}{c|}{215229}          & \multicolumn{1}{c|}{77485}              & \multicolumn{1}{c|}{2074704}                               & \multicolumn{1}{c|}{2570362}                  \\ \hline
\multicolumn{1}{|l|}{Avg summary length}              & \multicolumn{1}{c|}{16.08}          & \multicolumn{1}{c|}{22.82}           & \multicolumn{1}{c|}{26.05}              & \multicolumn{1}{c|}{35.29}                                 & \multicolumn{1}{l|}{}                         \\ \hline
\multicolumn{1}{|l|}{Avg input length}           & \multicolumn{1}{c|}{69.91}          & \multicolumn{1}{c|}{121.39}          & \multicolumn{1}{c|}{110.69}             & \multicolumn{1}{c|}{109.30}                                & \multicolumn{1}{l|}{}                         \\ \hline                         
\end{tabular}
%}
\caption{Full data statistics for \textit{VietMed-Sum}. Extention of Table \ref{data_stats}.}
\label{appx_data_stats}
\end{table}

\subsection{GPT Setup for Annotation}
\begin{figure}[ht]
  \centering
  \includegraphics[width=0.7\columnwidth]{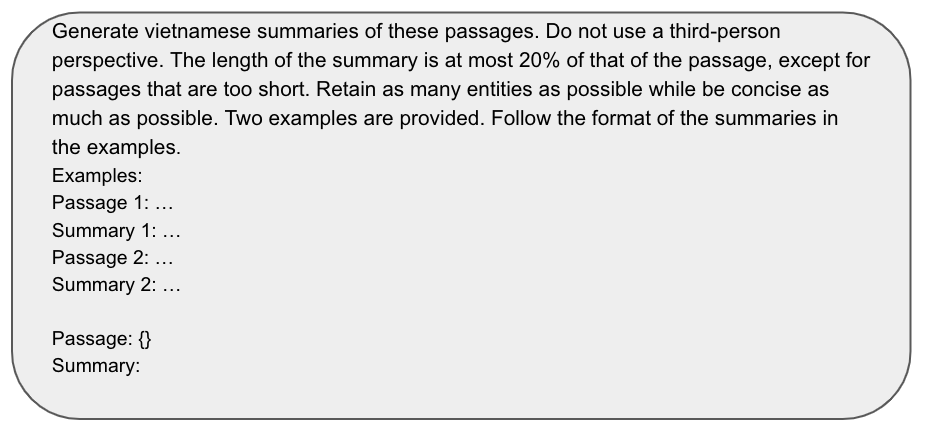}
  \caption{An example of prompt used for GPT summaries}
  \label{fig:prompt}
\end{figure}

We used GPT-3.5 Turbo with hyperparameters \textit{temperature}=0.7 and \textit{top}\_\textit{p}=0.9 to generate GPT summaries. 
Our choice to use GPT 3.5 Turbo for generating GPT summaries is motivated by balancing between the annotation cost and the overall data-model performance.
To improve the overall quality of these summaries, we used in-context learning with two examples. 
Figure \ref{fig:prompt} illustrates the prompt we used for in-context learning.

\subsection{Summarization Evaluation Metrics}
We used ROUGE \cite{lin2004rouge}, a metric commonly used for summarization, to evaluate our models. 
ROUGE measures the lexical overlap between the candidate and the reference summaries. 
We report the ROUGE-1, ROUGE-2, and ROUGE-L scores on the test set.
ROUGE-1 and ROUGE-2 measures the overlap between consecutive unigram and bigram, while ROUGE-L is based on the longest common subsequence.

\subsection{Detailed Information of Baseline Summarization Models}
Since some models lack the large version, our experiments were conducted on the base version. 
Fine-tuning was done using the Transformer library \cite{wolf2019huggingface}. 
\begin{itemize}
    \item \textbf{BARTpho} \cite{tran2022bartpho} is the Vietnamese BART \cite{lewis2019bart} pre-trained on 20GB of text from the Vietnamese Wikipedia and the Vietnamese news corpus. 
    We used both  $\text{BARTpho}_{\text{syllable}}$ and $\text{BARTpho}_{\text{word}}$, with number of parameters being 132M and 150M respectively. 
    \item \textbf{ViT5} \cite{phan2022vit5} is the Vietnamese T5 \cite{raffel2020exploring} pre-trained on 71GB of text from Vietnamese subset of CC100 \cite{conneau2019unsupervised}. 
    We included ViT5-vietnews, ViT5 finetuned on Vietnews \cite{8573420} for summarization, to observe if the previous fine-tuning knowledge could assist the model in our downstream task.
    The number of parameters is 310M. 
    \item \textbf{ViPubmedT5} \cite{phan2023enriching} is the ViT5 model pre-trained on 20GB of biomedical data from ViPubmed \cite{phan2023enriching}.
    % - the Vietnamese machine-translated version of Pubmed \cite{sen2008collective}. 
    The number of parameters is 220M.
    To integrate into our pipeline, we converted its t5x \cite{roberts2022t5x} checkpoint  to Flax \cite{flax2020github}.
    % \item \textbf{ViHealthBERT} is the first domain-specific pre-trained language model for Vietnamese healthcare. After initializing weights from PhoBERT, the model is trained on 25M health sentences mined from different sources.
\end{itemize}

\subsection{Details of Human Evaluation Experiments}
We used the human evaluation proposed by \cite{chen2021dialogsum_dataset, kryscinski2019human_eval_metric_sum}, which evaluates summaries based on 4 metrics:
\begin{itemize}
    \item Fluency: evaluates the quality of individual generated sentences
    \item Consistency: evaluates the factual alignment between the source text and generated summary
    \item Relevance: evaluates the importance of summary content
    \item Coherence: evaluates the collective quality of all sentences
\end{itemize}

We randomly picked 50 input text and their corresponding summaries in the test set, and ask a medical expert to give scores ranging from 1 to 5 based on 4 metrics mentioned above.
To remove human bias and subjectivity, the medical expert did not know which summary was generated by human annotators or by models.

\section{Additional Experimental Results}
\subsection{Additional Results for ASR Transcript Summarization}
\begin{table}[!ht]
\centering
%\resizebox{\columnwidth}{!}{%
\begin{tabular}{llcccc}
\toprule
\textbf{Data}&\textbf{Model} & \textbf{R-1} & \textbf{R-2} & \textbf{R-L} \\
\midrule
 \textbf{Global + Local}& $\text{BARTpho}_{\text{syllable}}$ & 57.19 & 34.27 & 44.97\\ 
 & $\text{BARTpho}_{\text{word}}$ & 55.69 & 33.66 & 43.95\\ 
 & ViT5- & 57.44 & 34.61 & 45.33 \\ 
 & ViT5-vietnews & 57.61 & 34.09 & 44.67  \\ 
 & ViPubmedT5 & 57.58 & 33.42 & 44.13\\ 
 \midrule
 \textbf{Local}& $\text{BARTpho}_{\text{syllable}}$ & 56.75 & 33.27 & 44.42 \\ 
 & $\text{BARTpho}_{\text{word}}$ & 55.22 & 32.15 & 43.05 \\ 
 & ViT5 & 54.41 & 30.69 & 41.91 \\ 
 & ViT5-vietnews & 56.39 & 32.60 & 43.35 \\ 
 & ViPubmedT5 & 54.21 & 30.15 & 41.04 \\ 
 \midrule
 \textbf{Global} & $\text{BARTpho}_{\text{syllable}}$ & 54.77 & 33.14 & 43.62 \\ 
 & $\text{BARTpho}_{\text{word}}$ & 52.96 & 32.22 & 42.02 \\ 
 & ViT5 & 55.92 & 34.11 & 44.57 \\
 & ViT5-vietnews & 55.95 & 33.87 & 44.25\\ 
 & ViPubmedT5 & 55.35 & 32.20 & 43.00 \\ 
 \midrule
 \textbf{SYN} & $\text{BARTpho}_{\text{syllable}}$ & 56.78 & 33.54 & 43.97 \\
 & $\text{BARTpho}_{\text{word}}$ & 56.08 & 32.92 & 43.19 \\ 
 & ViT5 & 57.32 & 34.70 & 44.71  \\ 
 & ViT5-vietnews & 56.73 & 33.45 & 44.08 \\ 
 & ViPubmedT5 & 56.62 & 34.01 & 44.21 \\ 
 \midrule
 \textbf{SYN} + \textbf{(Global + Local)}& $\text{BARTpho}_{\text{syllable}}$ & 56.58 & 34.32 & 44.69 \\ 
 & $\text{BARTpho}_{\text{word}}$ & 56.44 & 34.28 & 44.72 \\ 
 & ViT5 & 58.51 & 36.26 & 46.4  \\ 
 & ViT5-vietnews & 58.21 & 35.67 & 45.84 \\ 
 & ViPubmedT5 & 58.30 & 36.07 & 45.91 \\ 
 \midrule
 \textbf{SYN} $\rightarrow$ \textbf{(Global + Local)} & $\text{BARTpho}_{\text{syllable}}$ & 57.38 & 34.40 & 45.44 \\ 
 & $\text{BARTpho}_{\text{word}}$ & 56.42 & 34.24 & 44.65 \\ 
& ViT5 & 58.95 & 36.82 & 46.63 \\ 
& ViT5-vietnews & 58.22 & 35.34 & 46.02 \\ 
 & ViPubmedT5 & 57.58 & 33.42 & 44.13\\ 
\bottomrule
\end{tabular}
%}
\caption{Experimental results on ASR transcripts. We report the ROUGE on the \textit{GOLD} test global summaries. Extention of Table \ref{tab:asr_result}.}
\label{tab:new_vietmed_3_20k_label}
\end{table}

\begin{table}[!ht]
\centering

%\resizebox{\columnwidth}{!}{%
% \begin{tabular}{llccc}
% \toprule
% \textbf{Model} & \textbf{R-1} & \textbf{R-2} & \textbf{R-L} \\
\begin{tabular}{llccc}
\toprule
\textbf{Data}&\textbf{Model} & \textbf{R-1} & \textbf{R-2} & \textbf{R-L} \\

\midrule
\textbf{Gold} & $\text{BARTpho}_{\text{syllable}}$ & 55.32 & 32.46 & 44.04 \\
 & $\text{BARTpho}_{\text{word}}$ & 53.22 & 30.98 & 42.12 \\
 & ViT5 & 54.86 & 31.70 & 43.76 \\
 & ViT5-vietnews & 55.33 & 31.93 & 43.68 \\
 & ViPubmedT5 & 55.64 & 31.80 & 43.42 \\
\midrule
\textbf{Local} & $\text{BARTpho}_{\text{syllable}}$ & 54.44 & 31.41 & 43.16 \\
 & $\text{BARTpho}_{\text{word}}$ & 52.96 & 29.69 & 41.49 \\
 & ViT5 & 52.81 & 29.49 & 41.29 \\
 & ViT5-vietnews & 54.34 & 31.03 & 42.61 \\
 & ViPubmedT5 & 52.34 & 28.09 & 39.97 \\
\midrule
\textbf{Global} & $\text{BARTpho}_{\text{syllable}}$ & 53.69 & 31.75 & 43.01 \\
 & $\text{BARTpho}_{\text{word}}$ & 51.62 & 30.11 & 40.64 \\
 & ViT5 & 55.09 & 32.55 & 43.88 \\
 & ViT5-vietnews & 54.72 & 32.28 & 43.37 \\
 & ViPubmedT5 & 53.95 & 31.33 & 42.42 \\
\midrule
\textbf{SYN + (Global + Local)}& $\text{BARTpho}_{\text{syllable}}$ & 54.31 & 31.17 & 42.23 \\
 & $\text{BARTpho}_{\text{word}}$ & 53.76 & 31.18 & 41.92 \\
 & ViT5 & 55.21 & 32.47 & 43.29 \\
 & ViT5-vietnews & 54.01 & 31.49 & 41.85 \\
 & ViPubmedT5 & 54.28 & 31.00 & 42.41 \\
\midrule
\textbf{SYN} & $\text{BARTpho}_{\text{syllable}}$ & 54.78 & 31.92 & 43.46 \\
 & $\text{BARTpho}_{\text{word}}$ & 54.06 & 31.58 & 42.56 \\
 & ViT5 & 56.52 & 34.30 & 45.15 \\
 & ViT5-vietnews & 58.21 & 35.67 & 45.84 \\
 & ViPubmedT5 & 58.30 & 36.07 & 45.91 \\
 \midrule
\textbf{SYN} $\rightarrow$ \textbf{(Global + Local)} & $\text{BARTpho}_{\text{syllable}}$ & 55.46 & 32.57 & 44.38 \\
 & $\text{BARTpho}_{\text{word}}$ & 54.22 & 32.07 & 43.27 \\
& ViT5 & 56.78 & 34.43 & 45.52 \\
& ViT5-vietnews & 56.48 & 33.65 & 45.10 \\
 & ViPubmedT5-base & 55.48 & 31.57 & 43.06 \\
\bottomrule
\end{tabular}
%}

\caption{Experimental results on ASR transcripts. We report the ROUGE on the \textit{GOLD} test local summaries. Extention of Table \ref{tab:asr_result}.}
\label{tab:new_vietmed_3_20k_label_2}
\end{table}

Table \ref{tab:new_vietmed_3_20k_label} shows the baseline results on ASR transcript for global summaries.
Table \ref{tab:new_vietmed_3_20k_label_2} shows the baseline results on ASR transcript for local summaries.
Both tables are the extention of Table \ref{tab:asr_result}.

\subsection{Results for English-translated \textit{VietMed-Sum}}
To help international researchers, we translated our \textit{VietMed-Sum} dataset, called \textit{VietMed-Sum-en} and conducted experiments on it.
Table \ref{tab:comparison_en} shows the results of fine-tuning on \textit{VietMed-Sum-en} using some English pre-trained language models.

\begin{table}[!ht]
\centering
%\resizebox{\columnwidth}{!}{%
\begin{tabular}{l|l|ccc|ccc}
\hline
\textbf{Data} & \textbf{Model} & \multicolumn{3}{c|}{\textbf{Global Summaries}} & \multicolumn{3}{c}{\textbf{Local Summaries}} \\
&\textbf{Model} & \textbf{R-1} & \textbf{R-2} & \textbf{R-L} & \textbf{R-1} & \textbf{R-2} & \textbf{R-L} \\
\hline
\textbf{Gold}&$\text{BART-base}$ & 34.26 & 16.19 & 29.85 & 35.87 & 17.69 & 31.23 \\
&$\text{T5-base}$ & 33.49 & 14.73 & 28.89 & 34.67 & 15.93 & 29.88 \\
\hline
\textbf{Global}&$\text{BART-base}$ & 33.87 & 15.70 & 29.62 & 35.93 & 17.45 & 31.50 \\
&$\text{T5-base}$ & 30.69 & 13.49 & 26.29 & 31.65 & 14.47 & 27.04 \\
\hline
\textbf{Local}&$\text{BART-base}$ & 34.01 & 15.72 & 29.56 & 35.42 & 17.16 & 30.80 \\
&$\text{T5-base}$ & 33.05 & 14.59 & 28.18 & 34.37 & 15.77 & 29.33 \\
\hline
\end{tabular}%
%}
\caption{ROUGE scores of T5 and BART on the GOLD set of VietMed-Sum-en}
\label{tab:comparison_en}
\end{table}

\section{Data Annotation}
\subsection{Annotation Process and Data Quality Control}
We hosted meetings to discuss complex cases and create the annotation guideline with the help of a medical expert.
Two developers with basic medical training  independently edited GPT summaries based on the annotation guideline.
For each human-edited summary, the medical expert decided which version as the gold standard and requested re-annotation if there were any deviations from the annotation guideline.
For summaries not human-edited by the two developers, we called synthetic data.
Final annotation guideline is below.

\subsection{Annotation Guidelines}
We asked two annotators to follow the guideline described below:
\begin{enumerate}
    \item Keep the summary as short as possible without losing key information. 
    The length of the summary is at most 20\% of that of the passage (except too short dialogues).
    \item Retain as many medical named entities as possible as long as the limit is not exceeded.
    \item Retain the purpose of the passage, e.g. questions should be preserved as question summaries.
    \item Summaries must sound natural. 
\end{enumerate}

\section{Ethical Statements}
\subsection{Content Ownership}
According to OpenAI terms of use\footnote{https://openai.com/policies/terms-of-use}, we have our ownership of the content generated by ChatGPT: "As between you and OpenAI, and to the extent permitted by applicable law, you (a) retain your ownership rights in Input and (b) own the Output. 
We hereby assign to you all our right, title, and interest, if any, in and to Output."

Also according to Fair Use\footnote{https://www.copyright.gov/fair-use/}, we and our work are protected under Fair Use policy to conduct and publicly release research data for research purposes.
We state that we do not release data for commercial purposes, thus not rivaling any business parties.

\subsection{Privacy}
Understanding that medical data is sensitive, during the process of data generation and annotation, we carefully annonymized and removed any text that might reveal patient identities which might be incidently generated by LLMs.

\end{document}